\pdfoutput=1

\documentclass[11pt]{article}

\usepackage[]{acl}

\usepackage{times}
\usepackage{latexsym}

\usepackage[T1]{fontenc}

\usepackage[utf8]{inputenc}

\usepackage{microtype}

\usepackage{inconsolata}

\usepackage{authblk}
\usepackage{algorithm}
\usepackage{algorithmic}

\usepackage{amsmath}
\usepackage{times}
\usepackage{latexsym}
\usepackage{booktabs}
\usepackage{tabularx}
\usepackage{tikz}
\usepackage{pgfplots}
\usepackage{pgfplotstable}
\usepackage{enumitem}

\usepackage{soul}
\usepackage{pifont}
\usepackage{xcolor,colortbl}

\usepackage{cleveref}
\usepackage[]{mdframed}
\usepackage{framed}

\usepackage{url}
\usepackage{graphicx}

\usepackage{tabularx}
\usepackage{xcolor}
\usepackage{makecell}
\usepackage{enumitem}

\definecolor{Gray}{gray}{0.9}

\title{To Tell The Truth:\\Language of Deception and Language Models}

\author[$1$]{\textbf{Sanchaita Hazra}*}
\author[$2$]{\textbf{Bodhisattwa Prasad Majumder}*}
\affil[$1$]{University of Utah}
\affil[$2$]{Allen Institute for AI \protect \\ \texttt{sanchaita.hazra@utah.edu, bodhisattwam@allenai.org} \protect \\
\small \texttt{*contributes equally}}

\begin{document}
\maketitle
\begin{abstract}
Text-based false information permeates online discourses, yet evidence of people's ability to discern truth from such deceptive textual content is scarce. We analyze a novel TV game show data where conversations in a high-stake environment between individuals with conflicting objectives result in lies. We investigate the manifestation of potentially verifiable language cues of deception in the presence of objective truth, a distinguishing feature absent in previous text-based deception datasets. We show that there exists a class of detectors (algorithms) that have similar truth detection performance compared to human subjects, even when the former accesses only the language cues while the latter engages in conversations with complete access to all potential sources of cues (language and audio-visual). Our model, built on a large language model, employs a bottleneck framework to learn discernible cues to determine truth, an act of reasoning in which human subjects often perform poorly, even with incentives. Our model detects novel but accurate language cues in many cases where humans failed to detect deception, opening up the possibility of humans collaborating with algorithms and ameliorating their ability to detect the truth.
\end{abstract}

\section{Introduction}

Deception is pervasive in conversational dialogues. Individuals motivated by self-interest often feel compelled to embellish the truth to promote their interests at the expense of others. Misleading communication, such as false testimony \cite{tetterton2005using}, fake news \cite{shu2017fake}, identity fraud in dating sites \cite{lazarus2022exploring}, sock puppetry \cite{kumar2017army}, and propaganda campaigns \cite{allcott2017social}, abundant daily, impacts political, social, and economic outcomes. This exchange of information leading to the decision of who and what to believe necessitates the tacit development of truth detection capability during conversations \citep{Bond2006AccuracyOD}.


\begin{figure*}[t!]
\centering
\includegraphics[trim= 110 365 110 275,clip,width=0.95\linewidth]{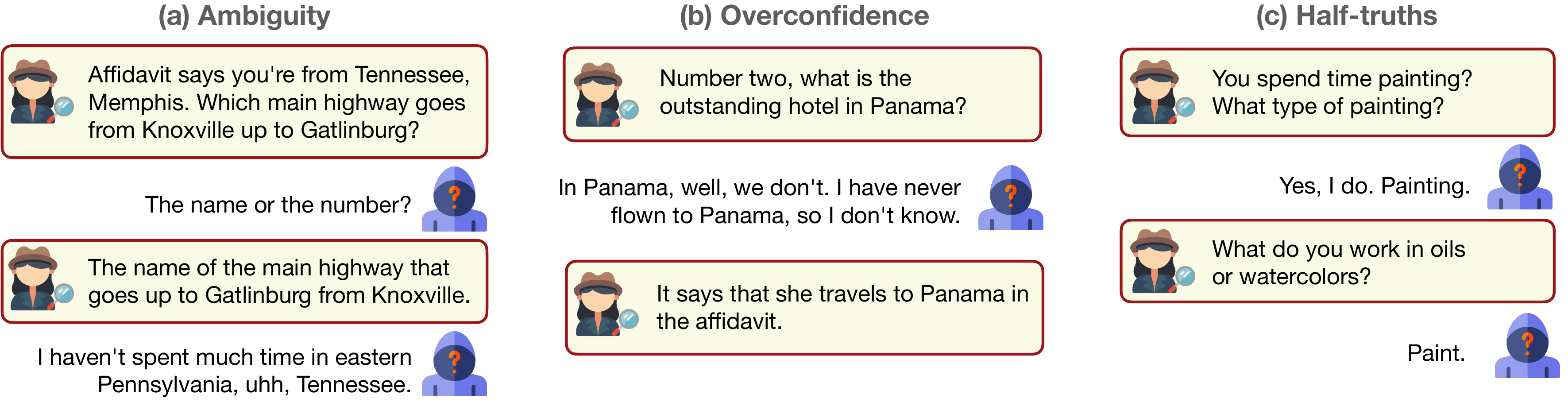}
\caption{Examples of language cues for detecting deception: ambiguity, overconfidence, and half-truths from our dataset (\textsc{T4Text}). When used features, they can significantly enhance detection ability for both models and humans.}
\label{fig:dataset}
\vspace{-0.5em}
\end{figure*}

In what follows, we explore if textual cues may increase the likelihood of fraud detection even in the presence of more overt visual or aural indicators. Consider the CEO scam, when fraudsters act as company executives to trick a victim into sending unauthorized wire transfers or divulging private information through email.
In addition, textual cues may be crucial for an impartial observer to identify duplicity in social media conversations when audio and visual cues are often manipulated with little to no chance of face-to-face conversations \cite{Rapoza_2021}.

This paper examines linguistic cues in a conversational exchange between contestants and judges participating in the TV game show, To Tell The Truth. In the game, the three contestants, under pretenses, mislead the four judges who attempt to infer the real central contestant (CC) via back-and-forth questioning. First, the game show offers a high-stake situation where contestants have a financial incentive to lie and deceive, and the judges are under pressure to perform in front of a crowd to detect the deception. Second, this data provides factual information to aid in assessing the contestant---a vital and distinguishing aspect of deception detection from the text than other datasets.


The study of deception detection using computational methods has traditionally focused on detecting `what' is the truth using multimodal cues \cite{soldner2019box}. The definition of truth is often convoluted and depends on context; hence computationally detecting `what is the truth' is challenging unless defined otherwise \cite{peskov2020takes}. Works that focused on exploring language cues \cite{fornaciari2013automatic,ott2013negative} mainly restricted their analysis to psycholinguistic and hand-engineered linguistic features, which may not extend to scenarios where such cues are missing. Even though psycholinguistic features indicate the interlocutor's intention on a syntactic or token level, they may not demonstrate a deeper semantic understanding of the text in the discourse context. Recent progress in language models' ability to understand text prompted us to benchmark large language models' (LLMs) performance for the first time in detecting deception.

Armed with two main questions: 1. Do enough language cues exist to discern truth from deceptive conversations without other multimodal cues; and 2. Can a class of algorithmic detectors identify these cues, compose them in a valid chain of reasoning, and identify the truth?---in this paper,
\textbf{we demonstrate a bottleneck framework} that progressively scans a deceptive conversation, analyzes each snippet by verifying utterances against objective truth, semantically understanding complex indicators of deception such as ambiguous responses, half-truths, and overconfidence, can satisfactorily reason its prediction for detecting deception. \textbf{We release a new conversational dataset}, To Tell The Truth from Text (\textsc{T4Text})\footnote{Code \& data: \url{https://github.com/sanchaitahazra/T4Text}}, unique to the previous datasets, that contains a verifiable objective truth, forming the basis of lie detection. Our model can detect deception in cases where all judges failed to detect lies correctly, indicating its ability to uncover new reasoning chains that might be insightful to humans to learn better predictors for deception. Our model sometimes fails, where judges could correctly identify deception, leaving room for researchers to advance the frontier of the model performance in deception detection.
\section{Related Work}

\paragraph{Deception detection.} Deception is an act emerging since the beginning of time with the Serpent and Eve in the Garden of Eden: \emph{And the serpent said unto the woman, Ye shall not surely die}. However, humans are often not very good at spotting these lies \cite{belot2012can,gneezy2005deception}, and are no better than making random decisions \cite{ockenfels2000experiment}. This raises the question of whether there are valid indicators of deception. According to studies \cite{wang2010pinocchio,depaulo2003cues,ekman1997deception,zuckerman1981verbal, wang2010pinocchio}, employing non-verbal signals such as visual, facial, and aural cues might significantly help distinguish sincere and opportunistic communication. 
Recent studies mention participants are significantly more accurate at spotting lies from both audio and videos (82\%) or only videos (66\%), compared to text (57\%) \cite{wittenberg2021minimal,groh2022human}. In this paper, we investigate if a computational model can instead detect deception in text using language cues.

\paragraph{Deception + NLP datasets.} Automated deception detection techniques so far predominantly utilized visual cues such as facial or eye movements to detect deception and time to response (\cite{meservy2005deception,gonzalez2019can}, or linguistic cues from transcriptions from court hearings \cite{fornaciari2013automatic}, deceptive hotel reviews \cite{ott2013negative}, news articles from Buzzfeed datasets \cite{potthast2017stylometric}, and fact-checked tweets \cite{van2022personal}. Datasets that require assigning specific individuals roles (lier/truth-teller) include a multimodal conversational dataset Box of Lies \cite{soldner2019box}, Golden Globes \cite{darai2013attraction} differ from Diplomacy gameplay \cite{peskov2020takes}, and Real or Spiel \cite{ho2016computer} where one can \emph{choose} to lie. In contrast, our derived dataset from this game show is conversational, grounded in a real deceptive environment with the presence of objective truth to detect the deception not present in existing datasets. The only work, \cite{banerjee2023experience} that investigated, To Tell The Truth gameshow neither built computational models nor analyzed cues from interactions. 

\paragraph{Models detecting deception.}
Computational models that focus on language cues mainly use psycholinguistic features \cite{Grlea2016PsycholinguisticFF, soldner2019box}, or syntactic parse of the texts \cite{soldner2019box} to identify predictors for deception. \citet{ho2016computer} utilized a power dynamics vocabulary to identify deception in long-term relationships. In our paper, we focus on more complex signals, such as ambiguity or half-truths, and benchmark the performance of LLMs for the first time in the context of deception detection from text.
\section{To Tell The Truth}
To examine conversational dynamics in deception detection, conversations among individuals laced with the intent to deceive given the prior unembellished truth is the key to our analysis. To Tell the Truth, season 1\footnote{This work was done under IRB\_00167477. ``To Tell The Truth'' videos used here were produced by CBS from 1956-59 and retrieved from YouTube. We consider using YouTube videos for research purposes to fall under the "fair use" clause, as stated: \url{https://www.youtube.com/intl/en-GB/yt/about/copyright/fair-use/}.}, is a game of deliberate misrepresentation and was aired on American TV weekly from 1956 to 1959. Every episode comprises multiple independent sessions, each lasting for about 8 minutes. A regular session comprises a host, four judges, and three contestants. One of the three contestants is the central contestant (CC), while the other two were imposters. 

\subsection{A game walk-through}
A session begins with the host asking every contestant's name (e.g., Jane Doe). The host publicly reads some factually true facts about the CC from a signed affidavit. The judges cross-question (for a fixed time) each contestant individually, by their respective numbers (Number one, two, or three).

The CC must answer truthfully to a question. The imposters, on the contrary, lie to impersonate the CC and deceive the judges. Following the definition of deception: ``Typically, when [someone] lies [they] say what [they] know to be false in an attempt to deceive the listener'' \cite{siegler1966lying}, our setup is a perfect case of deception. At the end of cross-questioning, judges disentangle the facts from the fiction and independently (and simultaneously) submit their votes for the real CC.

For each incorrect identification of the CC, the entire group of contestants was paid \$250 with a plausible individual maximum of \$333. Given the fundamental tenet of preferences, all contestants ideally desire increased collective financial gain. This translates to the intention of all, including the real CC, to deceive the judges successfully.




\subsection{Data Collection}
For this paper, we derive a slightly different game, To Tell The Truth \textbf{from Text} (\textsc{T4Text}). We transcribe 150 such games using the Whisper, a state-of-the-art transcription model with a word error rate of 8.81\% compared to human transcription's 7.61\% \cite{radford2023robust}.

During the early evaluation, we observed that all Whisper models (irrespective of size) often transcribed the proper names incorrectly. To address this, we manually review the automation-generated transcripts with the original video and corrected them for likely inconsistencies. Transcripts are cleaned for unnecessary noise or filler words in questions asked by judges (e.g., umm, uhh-hh) and any multi-lingual conversations beyond English. (``How do you pronounce your name in Russian?'', ``Please answer in French. I want to hear your accent.''). We do not include irrelevant mockery and conversations in-between judges or with the host. Owing to noise and inconsistency in the rationale for judges' votes, we have refrained from including them in our dataset.

\begin{figure*}[t!]
\centering
\includegraphics[trim= 165 330 100 260,clip,width=0.95\linewidth]{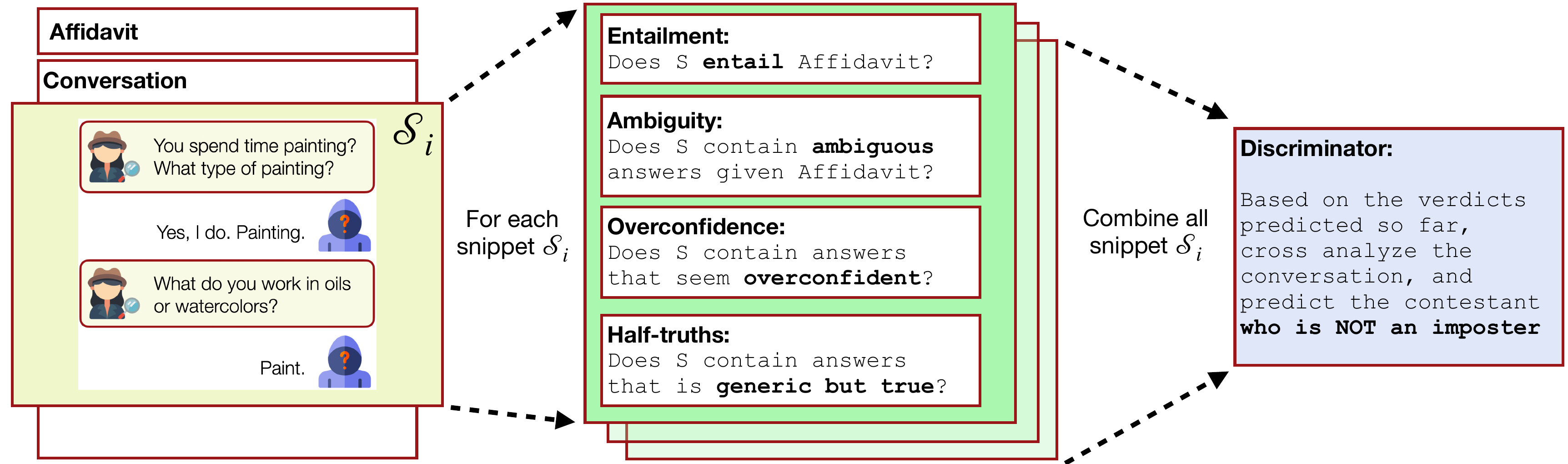}
\caption{Pipeline of the bottleneck model deriving bottleneck controls and the discriminator collates them for final prediction. We use few-shot LLMs to extract such controls that outperform an end-to-end approach.}
\label{fig:model}
\vspace{-0.5em}
\end{figure*}

\paragraph{Comment on data leak in LLMs.} Our dataset \emph{does not} exist in its textual form on the internet. Hence we do not necessarily run the risk of direct data contamination when applying LLM on them. 
However, for extra caution, we randomly swapped the contestant identities (e.g., changed number one to number three and vice-versa), which means it is not possible to ``copy'' the answer from the internet, if available, as the labels are now swapped too. Additionally, we replaced the participant names with placeholders (`Participant\_X'), where X is a random integer. 

\paragraph{Comment on selecting older sessions.} To Tell The Truth is a long-running TV show. We primarily consider the first season since not only has the show been intermittently revived from 2016-2022 on ABC, but the new episodes are also not as structured as the older ones. Few snippets of the full show that exist include more features of entertaining acts, unstructured questioning, and no compensation for the contestants (imposters), leading to uncertainties regarding the participants’ true intent to deceive.

\subsection{The {\textsc{T4Text}} Dataset}
\label{sec:dataset}
Each data point in \textsc{T4Text} entails three main components from an independent game session: the name of the real CC, the affidavit containing the objective truth about the CC, and the conversations (Q/As) between the judges and the contestants. Every data point, on average, has 12-15 Q/A pairs. 

Our dataset is novel since it is based on conversations around the 1950s before social media and the internet existed. Unique interpersonal interactions exist in \textsc{T4Text}, for instance, Edmund Hillary, the first person to summit Mount Everest in 1953, was one of the CCs in a session, but his appearance was unknown within the US entertainment industry, analogous to contemporary online crimes with unknown scammers.


\paragraph{Ambiguity/Randomness.}
According to \citet{ekman1997deception}, liars cannot keep their claims consistent, leading to ambiguity that exposes their lies. \Cref{fig:dataset} shows an example where the contestant mentions Pennsylvania while Tennessee is being discussed; this may have been an oversight on the contestant's part, or it may be a random interjection to stall a conversation; either way, it highlights the possibility that the contestant could be a liar. 

\paragraph{Overconfidence.} 
Overconfidence in the context of deception has been characterized in three ways: overestimating one's actual performance, overplacing one's performance in comparison to others, and excessive precision in one's beliefs \cite{moore2008trouble}. This behavior is similar to the contestants' intent to mislead the judges in \Cref{fig:dataset} where the contestant lied too confidently  \cite{serra2021mistakes} and hence made a small but important factual error for judges to understand that the individual is an imposter.

\paragraph{Half-truths.} In deceitful conversations, half-truths are less sinful than outright lies and explicit distortions \cite{carson2010lying}. Given the constraint of the real CC to answer truthfully, uttering half-truths becomes a prominent strategy to suppress facts vital for identification \cite{depaulo2003cues}. \Cref{fig:dataset} illustrates a contestant who does not mention what paint they use even after repeated questioning.

\paragraph{Dataset statistics.} \textsc{T4Text} is comprised of 150 data points with a volume of 86,746 words. There are 1546 utterances, including both judges and contestants as speakers. 450 unique contestants appeared in 150 sessions (datapoints), but judges reappeared from a unique set of size 56.



\section{Approach}
We investigate the possibility of using an LLM to uncover informative cues from the language of deception. Here we define the task, our base models, and a bottleneck model capable of reasoning the language cues for successful deception detection.

\subsection{Task}
We define the truth detection task as follows: given the name, affidavit ($\mathcal{A}$), and a conversation ($\mathcal{C}$) as input $x$, predict the real contestant as $y$ from an output label space of {\texttt{Number One}, \texttt{Number Two}, \texttt{Number Three}}. This is a discriminative task set up in the form of a 3-way classification problem. We use the terms \emph{deception detection} and \emph{truth identification} interchangeably in the rest of the paper.

\subsection{Base Models}
Our base model is an LLM.
We initiate the base model with a \textbf{task prompt} that includes the brief description and rules of \textsc{T4Text}
such as all contestants are incentivized to lie with the constraint that real CC still has to stick to the truth. The input followed by the task prompt contains the name of the CC, the affidavit, and the conversation between the judges and the contestants.
Please refer to the appendix for the complete prompt.

\subsection{Bottleneck Models}
Our base models process the input end-to-end to predict the real CC by performing necessary reasoning implicitly. However, implicit reasoning cannot be attributed to literature-backed linguistic cues, as discussed in \Cref{sec:dataset}. These cues can be explicitly extracted and used as a features. However, to ensure models restrict reasoning with these features, we use bottleneck models \cite{DBLP:conf/icml/KohNTMPKL20}. 

Identifying real CC involves assessing Q/A pairs addressed to a contestant at a time, as a \textbf{snippet} ($\mathcal{S}$) of the conversation and assessing the likelihood of the addressed contestant being the real CC. Our bottleneck models are employed through a set of bottleneck controls, which are the high-level predictors for deception detection (from \Cref{sec:dataset}). 

A bottleneck model takes the form of $f(g(\mathcal{S}))$; where $g$ is a mapping function that maps the input snippet $\mathcal{S}$ to a bottleneck control, predictive for deception and $f$ is the final discriminator that maps the intermediate bottleneck controls to the output label space. The success of the final prediction depends on the success of the intermediate functions generating bottleneck controls. We use LLMs for both $f$ and $g$. For each $g$, we write a bottleneck prompt ($p_{\text{bottleneck}}$) for each control, which we discuss here (also see \Cref{fig:model}). This simulates the systematic uncovering of these cues over the conversation, as the original judges would do.

\paragraph{Bottleneck controls.}
\begin{itemize}
    \item \textbf{Entailment:} As per game rules, each answer from the contestants should be verified in the light of the affidavit. We view this as an entailment task \cite{Tafjord2022EntailerAQ}. Given a premise and a hypothesis, an entailment task would be to predict if the hypothesis entails, contradicts, or does not relate to the premise. We set the affidavit $\mathcal{A}$ as the premise and a snippet $\mathcal{S}$ as the hypothesis and predict one of these: entail, contradiction, or neutral.
    \item \textbf{Ambiguity/Randomness:} Each snippet contains ambiguous or deliberately random responses from the contestants, indicative of deception (see \Cref{fig:dataset}). We develop a bottleneck prompt that takes a snippet $\mathcal{S}$ as an input to predict control values: ambiguous or unambiguous, in the light of the contestant being deceptive.  
    \item \textbf{Overconfidence:} Similarly as above, the next bottleneck prompt ascertains if the responses reveal overconfidence in a contestant (e.g., \Cref{fig:dataset}), indicating deception. The model predicts a verdict: overconfident or neutral.
    \item \textbf{Half-truths:} Finally, we develop a bottleneck prompt to decode an utterance as a half-truth (example in \Cref{fig:dataset}) to predict if the snippet contains half-truths and hence is indicative of deception or not.    
\end{itemize}

While the bottleneck controls are predicted for each conversation snippet, they can be derived either \emph{independently} or \emph{sequentially}. It is analogous to the original setting, where the snippets appear one by one, with the possibility that an older snippet may influence future questions from the judges and future answers from the contestants. For independent bottleneck controls, the mapping function is realized as $g(\mathcal{S}_i)$ for the $i$-th snippet. For sequential bottleneck control, the mapping function takes the form $g(\mathcal{S}_1,\cdots,\mathcal{S}_{i-1}, \mathcal{S}_i)$.

\paragraph{Discriminator.} The discriminator function $f({\cdot})$ takes annotated part of the conversations with the derived bottleneck controls for every snippet, to predict the real CC.

\paragraph{Hyperparameters.} We use OpenAI LLMs as the candidate base models and also for $f$ and $g$: \texttt{text-davinci-003} \cite{Brown2020LanguageMA}, \texttt{gpt3.5-turbo-16k}, and \texttt{gpt4} \cite{OpenAI2023GPT4TR}. For all GPT-3.5/4 experiments, we used \texttt{temperature} = 0, \texttt{max\_tokens} = 1024, and \texttt{top-p} = 1 (for nucleus sampling). The system prompt, the user prompt (for base models), and the bottleneck prompt are provided in \Cref{fig:system_prompt,fig:bottleneck_prompt}. We experimented with temperatures of 0.2 and 0.7, and the difference in the results was not statistically significant. Similarly, \texttt{top-p} = 0.95 did not yield any statistically significant different results. 

\subsection{Baselines and Evaluation}
One of the primary baselines for our system is to compare the model's performance with human performance. 

\paragraph{Base models.} For zero-shot models, our primary baselines will be the base models with all LLM variants that do not break the decision-making process through bottlenecks. \citet{Kojima2022LargeLM} show encouraging performance when a chain-of-thought (CoT) prompt is added to a zero-shot LLM: "\texttt{Let's think step by step},"---becomes our baseline.

\paragraph{Supervised Models.} For completeness, we also consider three supervised baselines where we only train the discriminator $f$ using XGBoost classifier, mirroring \cite{soldner2019box}. For the features required for the XGBoost classifier, we consider two options: psycholinguistic features from \citet{soldner2019box} and \texttt{gpt-3} embeddings of derived bottleneck controls from our \texttt{gpt-4}-based bottleneck model. For the LIWC-supervised baseline, we generate LIWC features for responses given by each contestant and concatenate them for the complete feature vector for the classifier. Finally, we train a BERT model \cite{DBLP:conf/naacl/DevlinCLT19}. For a fair comparison, we evaluate supervised models by a leave-one-out scheme spanning the full \textsc{T4Text} dataset.

\paragraph{Bottleneck Models.} For variations of our bottleneck approach, we create all possible combinations for $f$ and $g$ with our LLM variants. \texttt{gpt-4} as both $f(\cdot)$ and $f(\cdot)$ is our model, and rest 8 are baselines. We ablate four bottleneck controls individually while keeping the rest the same to compare with our model.
We also evaluate if independent or sequential bottleneck derivation affects model performance. Finally, \citet{soldner2019box} suggests LIWC features \cite{ott2013negative} are effective in predicting hidden intents in deceptive communication. We use such LIWC features as $g(\cdot)$, an alternative to our bottleneck features, pairing them with a \texttt{gpt-4} based discriminator. For all baselines, prompts are provided in the appendix.

\paragraph{Evaluation}
To evaluate model performance, we use accuracy and accuracy@2; the latter denotes if the correct prediction appears in the top two guesses. We use the session-level macro-average accuracy for human performance, as every session (one datapoint in \textsc{T4Text}) has predictions from 4 judges. We use a pairwise comparison in AMT and an absolute metric to evaluate the quality of the generated explanations from the models. For pairwise comparison, we measure \% of times explanations generated by our model are preferred by 3 human evaluators (in the majority) than explanations from a competing baseline. Following \citet{Majumder2021KnowledgeGroundedSV}, we use the e-ViL score on explanation where the models predicted accurately. We ask the annotators \emph{if an explanation is satisfactory} with four options: yes, partial-yes, partial-no, and no. This required us to take an intersection of samples when both comparing models generated correct predictions, an average of which was 31.

\begin{table}[t!]
\small
\centering
\resizebox{\linewidth}{!}{%
\begin{tabular}{@{}lcccc@{}}
\toprule
\bf Models & \bf Acc ($\uparrow$) & \bf Acc@2 ($\uparrow$) & \bf \% wins ($\uparrow$) & \bf $\kappa$ ($\uparrow$) \\ \midrule
Human* & 41.3 & -- & -- & -- \\ 
Random & 33.3 & 66.6 & -- & -- \\ \midrule
\rowcolor{Gray} \multicolumn{5}{@{}l}{\textbf{Base Models} (best setups)} \\ \midrule
GPT-3, 2-shot & 29.3 & 56.0 & 100 & 0.85\\
GPT-3.5, 2-shot & 33.3 & 70.0 & 97 & 0.81\\
GPT-4, 2-shot & 34.7 & 72.0 & 90 & 0.74\\ \midrule
\rowcolor{Gray} \multicolumn{5}{@{}l}{\textbf{CoT Models} (best setups)} \\ \midrule
GPT-3, 2-shot & 27.3 & 55.3 & 100 & 0.91\\
GPT-3.5, 2-shot & 30.0 & 65.3 & 100 & 0.91\\
GPT-4, 2-shot & 32.0 & 64.7 & 97 & 0.84\\ \midrule
\rowcolor{Gray} \multicolumn{5}{@{}l}{\textbf{Bottleneck Models} (best combinations)} \\ \midrule
$f$: GPT-3, $g$: GPT-4, 0-shot & 29.3 & 57.3 & 93 & 0.84\\
$f$: GPT-3, $g$: GPT-4, 2-shot & 30.0 & 58.1 & 93 & 0.85\\ \midrule
$f$: GPT-3.5, $g$: GPT-4, 0-shot & 35.3 & 71.3 & 77 & 0.70\\
$f$: GPT-3.5, $g$: GPT-4, 2-shot & 36.0 & 72.0 & 77 & 0.73\\ \midrule
$f$: GPT-4, $g$: LIWC & 33.3 & 71.3 & 100 & 0.91\\
$f$: GPT-4, $g$: GPT-4, 0-shot & \bf 39.3 & \bf 77.3 & 73 & 0.68\\
$f$: GPT-4, $g$: GPT-4, 2-shot & \bf 39.3 & \bf 77.3 & -- & --\\ \midrule
\rowcolor{Gray} \multicolumn{5}{@{}l}{\textbf{Supervised Models} ({leave-one-out})} \\ \midrule
BERT & 35.3 & 71.3 & -- & -- \\
GPT-3-emb. + XGBoost & 35.3 & 68.0 & -- & --\\
LIWC + XGBoost & 34.0 & 67.3 & -- & --\\
\bottomrule
\end{tabular}%
}
\caption{Accuracy (Acc) and Accuracy@2 (Acc@2) across models. \% wins indicate human pairwise evaluation for the explanations with Fleiss' $\kappa$ \cite{fleiss1973equivalence} scores as agreement.}
\label{tab:results}
\end{table}

\section{Experiments}
Our experiments are driven by the following hypotheses, and results are analyzed accordingly: 


\paragraph{Can few/zero-shot models detect deception?} \textbf{Yes}. \Cref{tab:results} contains the results. The bottleneck model with GPT-4 as both $g$ and $f$ performs the best in accuracy and accuracy@2. GPT-4 as $g$ consistently yields better performance than GPT-3.5 or GPT-3. Both GPT-3.5 and GPT-4 as $f$ are capable of detecting deception better than random. Base models perform worse than bottleneck models irrespective of the choice of the LLM and yield accuracy the same as random. However, the CoT models further decrease accuracy across all LLMs, primarily attributing to the CoT models' attempt to reason the potentially irrelevant or distracting information present in the conversation.

\begin{table}[t!]
\small
\centering
\resizebox{\linewidth}{!}{%
\begin{tabular}{@{}lcccc@{}}
\toprule
\bf Models & \bf Acc ($\uparrow$) & \bf Acc@2 ($\uparrow$) & \bf \% wins ($\uparrow$) & \bf $\kappa$ ($\uparrow$) \\ \midrule
Human* & 40.0 & -- & -- &  -- \\ 
Random & 33.3 & 66.6 & -- & -- \\ \midrule
\rowcolor{Gray} \multicolumn{5}{@{}l}{\textbf{Bottleneck Zero-shot Model. $f$: GPT-4, $g$: GPT-4, sequential}} \\ \midrule
Full model & 39 & 77 & -- & --\\
\hspace{0.5em} w/o entailment & 34 (-5) & 71 (-6) & 97 & 0.91\\
\hspace{0.5em} w/o ambiguity & 35 (-4) & 71 (-6) & 90 & 0.91\\
\hspace{0.5em} w/o overconfidence & 34 (-5) & 71 (-6) & 93 & 0.91\\ 
\hspace{0.5em} w/o half-truths & 32  (\textbf{-7}) & 68 (\textbf{-9}) & 100 & 0.96\\
independent-full & 36 (-3) & 75 (-2) & 93 & 0.85\\
\bottomrule
\end{tabular}%
}
\caption{Performance of the models under ablation. Numbers in parenthesis denotes performance drop from the `Full model'.}
\label{tab:ablation}
\end{table}

\paragraph{Why bottleneck models are better than base models?} \Cref{tab:ablation} shows the results of our ablation study that investigates the effect of each bottleneck control. The largest drop (7 points for Accuracy) occurs when the half-truth control is omitted, which aligns with our dataset analysis where real CC deliberately utters half-truths to deceive the judges, which, unless understood, may confuse the judges as well as models. We also find the quality of controls is better when derived sequentially (with conversation history) as compared to independent derivations, mirroring the original game setting.   

\paragraph{Are few-shot models better than zero-shot models?} \textbf{Depends}. For smaller or earlier LLMs (GPT-3, GPT-3.5), few-shot examples helps to improve their performance from zero-shot setup. However, for bottleneck models (and even in base/CoT models), GPT-4 achieves similar performance both in few-shot and zero-shot setups. Adding more demonstrations is often difficult due to LLM's limited context length. Further summarization \cite{Park2023GenerativeAI} or selection of few-shot examples \cite{Madaan2022MemoryassistedPE} are possible, but we leave it as a future work. 

\paragraph{Are supervised models better than few/zero-shot models?} \textbf{No}. Despite training on almost the full dataset (leave-one-out), the state-of-the-art supervised models do not outperform our zero-shot bottleneck models, indicating the superior ability of the LLMs to derive better bottleneck controls and act as a better discriminator.    

\begin{figure}[t!]
\centering
\includegraphics[trim= 175 165 380 295,clip,width=\linewidth]{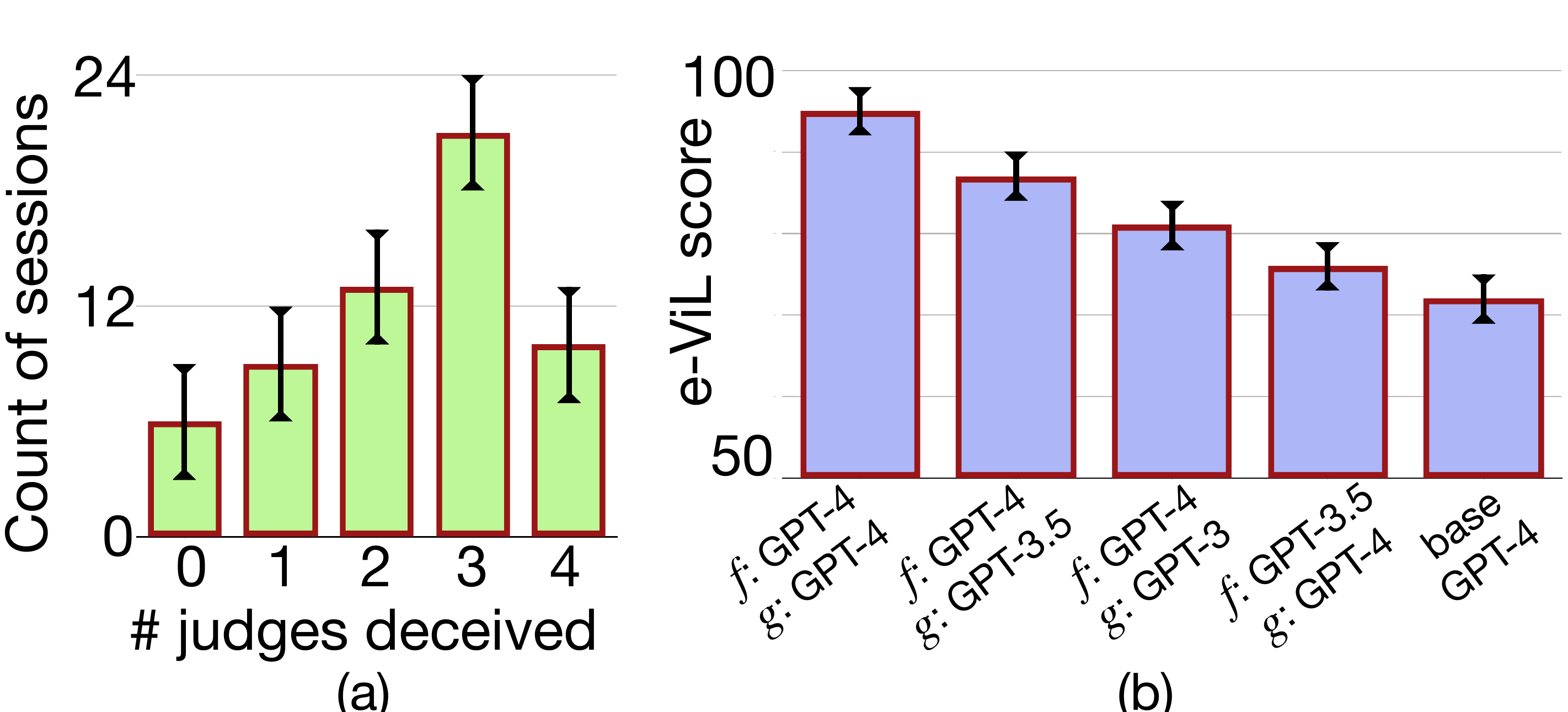}
\caption{\textbf{(a)} Histogram of a number of judges who were deceived in session where our model predicted correctly. The distribution is skewed; skewness = $-0.501$, indicating our model predicts correctly, significantly ($p < 0.1$) more in sessions where more judges were deceived. \textbf{(b)} e-ViL scores for explanations from our top-5 accurate models, showing our best model generates more satisfactory explanations.}
\label{fig:barplots}
\end{figure}

\paragraph{Can models explain their chain-of-reasoning to detect deception?} \textbf{Yes}. \Cref{tab:results} shows results for human evaluation. In the pairwise comparison, our best model (bottleneck, $f$: GPT-4, $g$: GPT-4) wins unanimously against all other competing baselines. Indeed, the quality of the explanations is distinctly worse for bottleneck models using GPT-3 or GPT-3.5 when compared to GPT-4, denoting GPT-4's ability to better bottleneck controls, which form the basis of a good explanation. Similarly, base models cannot generate high-quality explanations mainly due to their unconstrained nature of generation. \Cref{fig:barplots}b shows that our model has the highest e-ViL score, reflecting the trend from pairwise comparisons. 

\paragraph{Do self-consistency improve reasoning performance?} \textbf{Not significantly.} \citet{DBLP:conf/iclr/0002WSLCNCZ23} showed adding self-consistency prompt improves few-shot models reasoning performance. We implemented self-consistency in our best bottleneck model as well as in the COT baseline. Self-consistency improved the performance of the CoT baseline. However, it marginally improved the performance of our bottleneck models (not statistically significant). This indicates that despite using sophisticated task-specific prompts and refinement techniques, the dataset still remains a challenging case for state-of-the-art techniques involving LLMs. \Cref{tab:selfconsistency} shows the results of the self-consistency study for the CoT baseline and our bottleneck model (all with GPT-4).

\begin{table}[h!]
\small
\resizebox{\linewidth}{!}{%
\begin{tabular}{lcc}
\toprule
\textbf{Model}                   & \textbf{Acc ($\uparrow$)}  & \textbf{Acc@2 ($\uparrow$)} \\ \midrule
CoT                     & 32.0 & 64.7  \\
CoT + Self-consistency  & 33.3 (+1.3) & 68.0 (+3.3) \\
Ours                    & 39.3 (+7.3) & 77.3 (+12.6) \\
Ours + Self-consistency & 40.0 (+8) & 77.3 (+12.6) \\
\bottomrule
\end{tabular}%
}
\caption{Performance of the models with self-consistency. Numbers in parenthesis denote performance increase from the `CoT model.'}
\label{tab:selfconsistency}
\end{table}

\paragraph{Are there any potential variations in performance across different language models?} The choice of our language model (GPT-4 ) may significantly impact the results. A more comprehensive analysis of various models could strengthen the study. We extend our experiments by adding results from two more language models: Llama-2 \cite{Touvron2023Llama2O} and Bard\footnote{\url{https://gemini.google.com/}}. Llama-2 is an open-source language model. While Bard (now Gemini) has access to Google Search and YouTube, the measures taken to prevent data leaks make Bard a fair candidate for a competing baseline. \Cref{tab:variation} shows the performance of GPT-4 dominating over other language models.

\begin{table}[h!]
\small
\resizebox{\linewidth}{!}{%
\begin{tabular}{lcc}
\toprule
\textbf{Model}             & \textbf{Acc ($\uparrow$)}  & \textbf{Acc@2 ($\uparrow$)} \\ \midrule
GPT-4 (2-shot)             & \textbf{34.7} & \textbf{72.0}  \\
Llama-2 (2-shot)           & 32.0 (-2.7)   & 64.7 (-7.3)    \\
Bard (2-shot)              & 32.7 (-2.1)   & 65.3 (-6.7)    \\ \midrule
\textbf{Bottleneck Models} &               &                \\ \midrule
GPT-4 (\textit{f}, \textit{g}, 2-shot)       & \textbf{39.3} & \textbf{77.3}  \\
Llama-2 (\textit{f}, \textit{g}, 2-shot)     & 35.3 (-3)     & 71.3 (-6)      \\
Bard (\textit{f}, \textit{g}, 2-shot)        & 36.0 (-3.3)   & 72.0 (-5.3)    \\
\bottomrule
\end{tabular}%
}
\caption{Performance across different language models.}
\label{tab:variation}
\end{table}

\begin{figure*}[t!]
\centering
\includegraphics[trim= 50 383 50 43,clip,width=\linewidth]{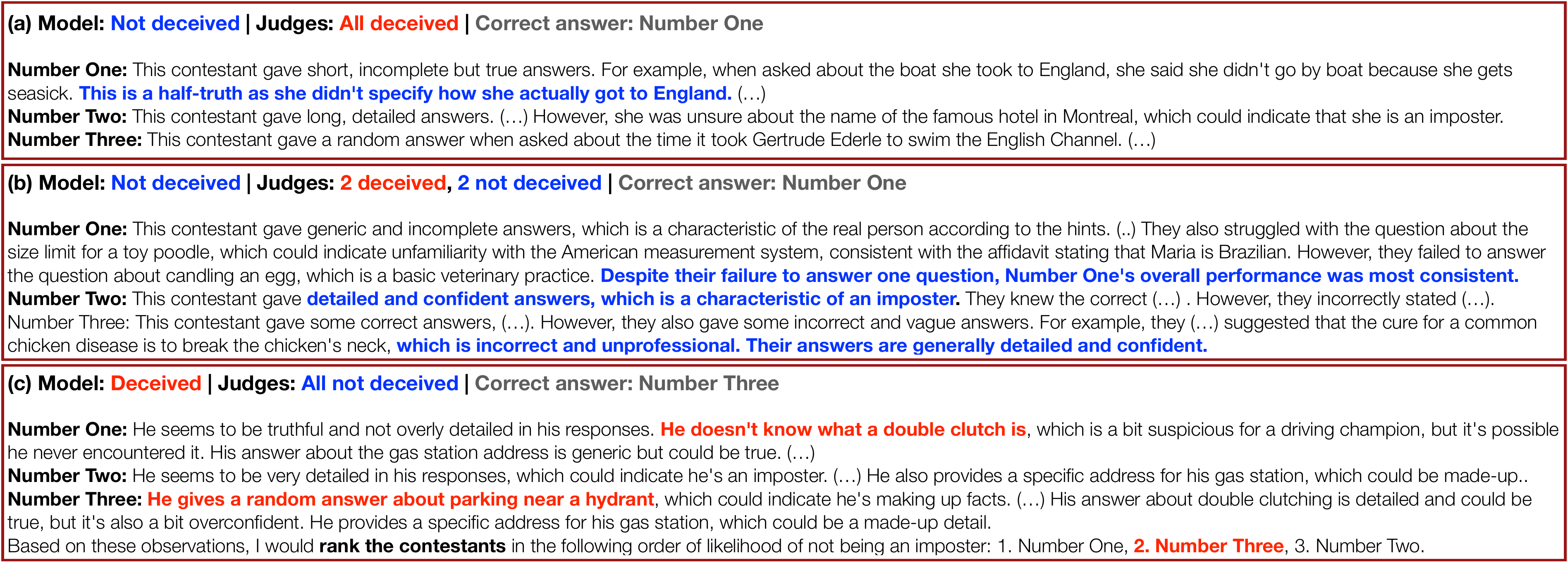}
\caption{Model generated explanations. Correct cues are colored in \textcolor{blue}{blue}, and wrong cues are colored in \textcolor{red}{red}.}
\label{fig:explanations}
\vspace{-0.5em}
\end{figure*}

\paragraph{Qualitative analysis of the model generated explanations.}
\Cref{fig:explanations} exhibits model-generated examples. In \Cref{fig:explanations}a, the model correctly identifies the CC, whereas \emph{All} of the judges fail. They mistook detailed information and deceptive appearances of the imposters (here, a swimmer's tan, athletic body, etc.). The model identifies bizarre details and randomness, even in the incorrect information by the imposters, while trying to capture the essence of the half-truths (highlighted) given by the real CC. In similar other cases where the model outperforms all the judges, the model mentions humor, overconfidence, deliberate incompleteness in answers, and wordy descriptions as possible cues of deceptive contestants. 

In \Cref{fig:explanations}b, the model predicts CC correctly, and the success for the judges is divided into half. Here, the model accurately recognizes a mistakenly erroneous response provided by the real CC as well as an ``overall consistent behavior,'' denoting honesty. The judges who were not duped might have noticed CC's constant behavior or the inconsistencies among the imposters, while the duped judges might have relied on CC's inaccurate response to draw their conclusions. One of the judges who was not deceived in this case reappeared in the majority of the sessions, which indicates possible learning to support her claim \cite{banerjee2023experience}.

\paragraph{Can models detect deception better than humans?} \textbf{Depends}. Note that the human judges had access to all potential audio-visual and language cues in the original setting. Audio-visual cues are very important, when present, in deception detection primarily due to the ease and abundance of detecting them \cite{depaulo2003cues}. Indeed, judges consistently use visual cues to explain their guesses, such as detecting the real swimmer/climber observing sun tan on the contestants or picking on the twinkle in the eyes for detecting a real joker. Considering this, our best model's performance is almost comparable with human performance---in absolute terms, our model incorrectly predicted 3/150 more examples compared to human judges. This is slightly different than results obtained in \cite{soldner2019box}, where they show linguistic features are the weakest predictors. However, it might be due to the fact that their linguistic features (part of them are LIWC features) were not expressive enough, or our dataset contains more linguistic cues than theirs. This also signifies the uniqueness of \textsc{T4Text} with the existence of the objective truth (affidavit), the basis of deception detection, and cross-questions revealing important factual and language cues to detect the real CC.

\paragraph{Do models detect deception for a complementary set than humans?} \textbf{Yes}. Among the 59/150 correct predictions by our best model, we analyze how human judges performed in those game sessions. \Cref{fig:barplots}a shows the histogram of the discrete variable--the number of judges deceived per session, considering only 59 sessions said above. The plot is skewed towards a higher number of deceived judges per session, indicating that our model performed better on a set where humans detect deception poorly. This implies our model discovered novel reasoning pathways, purely based on language cues, for correct predictions, which human judges failed at, even with the presence of multimodal cues and incentives. We qualitatively examine the explanations generated by the model where all judges got deceived (10/59) and discovered that our model generated more informative and plausible reasoning chains than explanations by human judges mentioned during the original sessions (see \Cref{fig:explanations}).

\paragraph{Can bottleneck features enhance human detection rate?} \textbf{Yes}. We performed a correlation analysis between the presence of a bottleneck feature and human prediction jointly using a multivariate regression model on cases where having these features significantly helped the computational model to predict correctly. On the intersection where humans were wrong but the bottleneck model was correct, we find either no correlation ($p < 0.05$ for entailment and ambiguity) or negative correlation ($p < 0.05$ for overconfidence and half-truths), indicating humans when poor at detecting such linguistic cues suffers significantly in detecting lies.

\paragraph{Are some sessions more difficult than others?} \textbf{To the model, yes; to humans, mostly no}. \citet{banerjee2023experience} confirm no selection bias among the contestants across the sessions in the original game, indicating similar difficulty levels for the judges across sessions. They also highlight weakly significant selection bias among the judges, indicating better performance via learning and possible further selection as a judge in the session might have influenced their performance. Qualitatively, we see models' errors clustered around sessions where judges use non-language cues successfully; however, model predictions among themselves were highly correlated ($\rho = 0.7, p < 0.05$) for each underlying LLM variant.  
\Cref{fig:explanations}c shows a case where the model fails to recognize the CC, but none of the judges were deceived. The model hallucinates in its reasoning to decide between Number One and Number Three as CC. It provides reasons for both to have incomplete, somewhat more detailed, a few factually accurate responses, and however forth—making the prediction at random. However, the real CC exists in the top two choices from the models, indicating judges picked up cues beyond language. Indeed, in this case, judges recognized the Southern accent of the CC and verified via the affidavit, a critical aural cue that was not available to our model.

\section{Conclusion and Outlook}

In this paper, we first showed the existence of a class of algorithmic detectors based on LLMs that can successfully identify language cues of deception without the presence of other visual or audio cues.  We contribute a novel dataset \textsc{T4Text} for deception detection in the presence of objective truth and achieve a model performance comparable to human performance. We further find that our best model performs well in cases where humans perform poorly and discover novel language models that could augment human reasoning to detect deception, opening up the possibility of human-LLM collaborations to combat misinformation.  

This paper advocates for human-AI collaboration, emphasizing the need for additional evidence on human dependence on algorithms in detecting textual deception. Using the methodology of incentivized decision-making from behavioral economics, our ongoing work seeks to observe and categorically elucidate human reliance on our AI model in different conditions (e.g., black box, glass box). Obtaining predictive features from our best model and those that drive human performance can provide us with a comprehensive understanding of why humans perform poorly in text-based deception detection tasks and whether AI tools can aid humans in such a crucial task that has lasting social, economic, and ecological impacts.


\section{Limitations}
We acknowledge that online misinformation can be very different in nature than lies in our dataset; however, we find examples of false information in Quora, Reddit where non-experts with propaganda use strategies like half-truths to misguide people. \textsc{T4Text} is a relatively small dataset; however, we showed that statistically significant analysis can be done with it. The human prediction data is derived from the original game show; hence the setup may not match exactly when we are evaluating text models. We are running additional human experiments to gather true human performance on \textsc{\textsc{T4Text}}. 

\section{Ethical Concerns} 
The dataset is in English. The original sessions occurred in the 1950s; hence we do not observe an equitable diversity in gender when it comes to the gender of the contestants. For all sessions, there were two female judges and two male judges. We occasionally observe judges asking questions that are biased toward gender or race; hence, any model that will be trained on this dataset may risk containing similar bias. In our paper, we do not train any generative model on this data, minimizing that risk. We acknowledge the potential misuse of such truth-detection systems, and we are following up with controlled experiments to understand if humans would over-rely on such systems.

\paragraph{Full declaration on dataset use.} This work was done under \texttt{IRB\_00167477}, which was approved for using freely available ``To Tell The Truth'' videos for transcription (we use open-source models) and evaluating humans' and algorithms' ability to detect deception from text.
``To Tell The Truth'' videos used in this paper were produced by CBS from 1956-59 and retrieved from YouTube. We consider using YouTube videos for research purposes to fall under the ``fair use'' clause, as stated: \url{https://www.youtube.com/intl/en-GB/yt/about/copyright/fair-use/}. We do not use any other data except videos, especially anything that is produced by YouTube.
We are additionally inspired by \citet{soldner2019box}, who use similar YouTube videos (of a different gameshow) for detecting deception from multimodal cues.
We release the video transcripts, the only data used in our paper, under a Creative Commons license (CC BY 4.0 DEED).

\section*{Acknowledgements} We sincerely thank Haimanti Bhattacharya and Subhasish Dugar for improving the foundational idea of this work. We also thank Chris Callison-Burch, Tushar Khot, and Peter Clark from the Allen Institute for AI for their generous feedback on the conclusions presented in this work.

\bibliography{anthology,custom}

\begin{thebibliography}{45}
\expandafter\ifx\csname natexlab\endcsname\relax\def\natexlab#1{#1}\fi

\bibitem[{Allcott and Gentzkow(2017)}]{allcott2017social}
Hunt Allcott and Matthew Gentzkow. 2017.
\newblock Social media and fake news in the 2016 election.
\newblock \emph{Journal of economic perspectives}, 31(2):211--236.

\bibitem[{Banerjee et~al.(2023)Banerjee, Ghosh, and Hazra}]{banerjee2023experience}
Priyodorshi Banerjee, Sanmitra Ghosh, and Sanchaita Hazra. 2023.
\newblock Experience, learning and the detection of deception.
\newblock \emph{Journal of Economic Criminology}, page 100010.

\bibitem[{Belot et~al.(2012)Belot, Bhaskar, and Van De~Ven}]{belot2012can}
Mich{\`e}le Belot, V~Bhaskar, and Jeroen Van De~Ven. 2012.
\newblock Can observers predict trustworthiness?
\newblock \emph{Review of Economics and Statistics}, 94(1):246--259.

\bibitem[{Bond and DePaulo(2006)}]{Bond2006AccuracyOD}
Charles F.~Jr. Bond and Bella DePaulo. 2006.
\newblock Accuracy of deception judgments.
\newblock \emph{Personality and Social Psychology Review}, 10:214 -- 234.

\bibitem[{Brown et~al.(2020)Brown, Mann, Ryder, Subbiah, Kaplan, Dhariwal, Neelakantan, Shyam, Sastry, Askell, Agarwal, Herbert-Voss, Krueger, Henighan, Child, Ramesh, Ziegler, Wu, Winter, Hesse, Chen, Sigler, Litwin, Gray, Chess, Clark, Berner, McCandlish, Radford, Sutskever, and Amodei}]{Brown2020LanguageMA}
Tom~B. Brown, Benjamin Mann, Nick Ryder, Melanie Subbiah, Jared Kaplan, Prafulla Dhariwal, Arvind Neelakantan, Pranav Shyam, Girish Sastry, Amanda Askell, Sandhini Agarwal, Ariel Herbert-Voss, Gretchen Krueger, T.~J. Henighan, Rewon Child, Aditya Ramesh, Daniel~M. Ziegler, Jeff Wu, Clemens Winter, Christopher Hesse, Mark Chen, Eric Sigler, Mateusz Litwin, Scott Gray, Benjamin Chess, Jack Clark, Christopher Berner, Sam McCandlish, Alec Radford, Ilya Sutskever, and Dario Amodei. 2020.
\newblock \href {https://api.semanticscholar.org/CorpusID:218971783} {Language models are few-shot learners}.
\newblock \emph{ArXiv}, abs/2005.14165.

\bibitem[{Carson(2010)}]{carson2010lying}
Thomas~L Carson. 2010.
\newblock \emph{Lying and deception: Theory and practice}.
\newblock OUP Oxford.

\bibitem[{Darai and Gr{\"a}tz(2013)}]{darai2013attraction}
Donja Darai and Silvia Gr{\"a}tz. 2013.
\newblock Attraction and cooperative behavior.
\newblock \emph{University of Zurich Department of Economics Working Paper}, (82).

\bibitem[{DePaulo et~al.(2003)DePaulo, Lindsay, Malone, Muhlenbruck, Charlton, and Cooper}]{depaulo2003cues}
Bella~M DePaulo, James~J Lindsay, Brian~E Malone, Laura Muhlenbruck, Kelly Charlton, and Harris Cooper. 2003.
\newblock Cues to deception.
\newblock \emph{Psychological bulletin}, 129(1):74.

\bibitem[{Devlin et~al.(2019)Devlin, Chang, Lee, and Toutanova}]{DBLP:conf/naacl/DevlinCLT19}
Jacob Devlin, Ming{-}Wei Chang, Kenton Lee, and Kristina Toutanova. 2019.
\newblock \href {https://doi.org/10.18653/V1/N19-1423} {{BERT:} pre-training of deep bidirectional transformers for language understanding}.
\newblock In \emph{{NAACL-HLT}}.

\bibitem[{Ekman(1997)}]{ekman1997deception}
Paul Ekman. 1997.
\newblock Deception, lying, and demeanor.
\newblock \emph{States of mind: American and post-Soviet perspectives on contemporary issues in psychology}, pages 93--105.

\bibitem[{Fleiss and Cohen(1973)}]{fleiss1973equivalence}
Joseph~L Fleiss and Jacob Cohen. 1973.
\newblock The equivalence of weighted kappa and the intraclass correlation coefficient as measures of reliability.
\newblock \emph{Educational and psychological measurement}, 33(3):613--619.

\bibitem[{Fornaciari and Poesio(2013)}]{fornaciari2013automatic}
Tommaso Fornaciari and Massimo Poesio. 2013.
\newblock Automatic deception detection in italian court cases.
\newblock \emph{Artificial intelligence and law}, 21:303--340.

\bibitem[{G{\^i}rlea et~al.(2016)G{\^i}rlea, Girju, and Amir}]{Grlea2016PsycholinguisticFF}
Codruta G{\^i}rlea, Roxana Girju, and Eyal Amir. 2016.
\newblock \href {https://api.semanticscholar.org/CorpusID:12091113} {Psycholinguistic features for deceptive role detection in werewolf}.
\newblock In \emph{North American Chapter of the Association for Computational Linguistics}.

\bibitem[{Gneezy(2005)}]{gneezy2005deception}
Uri Gneezy. 2005.
\newblock Deception: The role of consequences.
\newblock \emph{American Economic Review}, 95(1):384--394.

\bibitem[{Gonzalez-Billandon et~al.(2019)Gonzalez-Billandon, Aroyo, Tonelli, Pasquali, Sciutti, Gori, Sandini, and Rea}]{gonzalez2019can}
Jonas Gonzalez-Billandon, Alexander~M Aroyo, Alessia Tonelli, Dario Pasquali, Alessandra Sciutti, Monica Gori, Giulio Sandini, and Francesco Rea. 2019.
\newblock Can a robot catch you lying? a machine learning system to detect lies during interactions.
\newblock \emph{Frontiers in Robotics and AI}, 6:64.

\bibitem[{Groh et~al.(2022)Groh, Sankaranarayanan, Lippman, and Picard}]{groh2022human}
Matthew Groh, Aruna Sankaranarayanan, Andrew Lippman, and Rosalind Picard. 2022.
\newblock Human detection of political deepfakes across transcripts, audio, and video.
\newblock \emph{arXiv preprint arXiv:2202.12883}.

\bibitem[{Ho et~al.(2016)Ho, Hancock, Booth, and Liu}]{ho2016computer}
Shuyuan~Mary Ho, Jeffrey~T Hancock, Cheryl Booth, and Xiuwen Liu. 2016.
\newblock Computer-mediated deception: Strategies revealed by language-action cues in spontaneous communication.
\newblock \emph{Journal of Management Information Systems}, 33(2):393--420.

\bibitem[{Koh et~al.(2020)Koh, Nguyen, Tang, Mussmann, Pierson, Kim, and Liang}]{DBLP:conf/icml/KohNTMPKL20}
Pang~Wei Koh, Thao Nguyen, Yew~Siang Tang, Stephen Mussmann, Emma Pierson, Been Kim, and Percy Liang. 2020.
\newblock \href {http://proceedings.mlr.press/v119/koh20a.html} {Concept bottleneck models}.
\newblock In \emph{ICML}.

\bibitem[{Kojima et~al.(2022)Kojima, Gu, Reid, Matsuo, and Iwasawa}]{Kojima2022LargeLM}
Takeshi Kojima, Shixiang~Shane Gu, Machel Reid, Yutaka Matsuo, and Yusuke Iwasawa. 2022.
\newblock \href {https://api.semanticscholar.org/CorpusID:249017743} {Large language models are zero-shot reasoners}.
\newblock \emph{ArXiv}, abs/2205.11916.

\bibitem[{Kumar et~al.(2017)Kumar, Cheng, Leskovec, and Subrahmanian}]{kumar2017army}
Srijan Kumar, Justin Cheng, Jure Leskovec, and VS~Subrahmanian. 2017.
\newblock An army of me: Sockpuppets in online discussion communities.
\newblock In \emph{Proceedings of the 26th international conference on world wide web}, pages 857--866.

\bibitem[{Lazarus et~al.(2022)Lazarus, Button, and Kapend}]{lazarus2022exploring}
Suleman Lazarus, Mark Button, and Richard Kapend. 2022.
\newblock Exploring the value of feminist theory in understanding digital crimes: Gender and cybercrime types.
\newblock \emph{The Howard Journal of Crime and Justice}, 61(3):381--398.

\bibitem[{Madaan et~al.(2022)Madaan, Tandon, Clark, and Yang}]{Madaan2022MemoryassistedPE}
Aman Madaan, Niket Tandon, Peter Clark, and Yiming Yang. 2022.
\newblock \href {https://api.semanticscholar.org/CorpusID:246016194} {Memory-assisted prompt editing to improve gpt-3 after deployment}.
\newblock \emph{ArXiv}, abs/2201.06009.

\bibitem[{Majumder et~al.(2021)Majumder, Camburu, Lukasiewicz, and McAuley}]{Majumder2021KnowledgeGroundedSV}
Bodhisattwa~Prasad Majumder, Oana-Maria Camburu, Thomas Lukasiewicz, and Julian McAuley. 2021.
\newblock \href {https://api.semanticscholar.org/CorpusID:248963782} {Knowledge-grounded self-rationalization via extractive and natural language explanations}.
\newblock In \emph{International Conference on Machine Learning}.

\bibitem[{Meservy et~al.(2005)Meservy, Jensen, Kruse, Burgoon, Nunamaker, Twitchell, Tsechpenakis, and Metaxas}]{meservy2005deception}
Thomas~O Meservy, Matthew~L Jensen, John Kruse, Judee~K Burgoon, Jay~F Nunamaker, Douglas~P Twitchell, Gabriel Tsechpenakis, and Dimitris~N Metaxas. 2005.
\newblock Deception detection through automatic, unobtrusive analysis of nonverbal behavior.
\newblock \emph{IEEE Intelligent Systems}, 20(5):36--43.

\bibitem[{Moore and Healy(2008)}]{moore2008trouble}
Don~A Moore and Paul~J Healy. 2008.
\newblock The trouble with overconfidence.
\newblock \emph{Psychological review}, 115(2):502.

\bibitem[{Ockenfels and Selten(2000)}]{ockenfels2000experiment}
Axel Ockenfels and Reinhard Selten. 2000.
\newblock An experiment on the hypothesis of involuntary truth-signalling in bargaining.
\newblock \emph{Games and Economic Behavior}, 33(1):90--116.

\bibitem[{OpenAI(2023)}]{OpenAI2023GPT4TR}
OpenAI. 2023.
\newblock \href {https://api.semanticscholar.org/CorpusID:257532815} {Gpt-4 technical report}.
\newblock \emph{ArXiv}, abs/2303.08774.

\bibitem[{Ott et~al.(2013)Ott, Cardie, and Hancock}]{ott2013negative}
Myle Ott, Claire Cardie, and Jeffrey~T Hancock. 2013.
\newblock Negative deceptive opinion spam.
\newblock In \emph{Proceedings of the 2013 conference of the north american chapter of the association for computational linguistics: human language technologies}, pages 497--501.

\bibitem[{Park et~al.(2023)Park, O'Brien, Cai, Morris, Liang, and Bernstein}]{Park2023GenerativeAI}
Joon~Sung Park, Joseph~C. O'Brien, Carrie~J. Cai, Meredith~Ringel Morris, Percy Liang, and Michael~S. Bernstein. 2023.
\newblock \href {https://api.semanticscholar.org/CorpusID:258040990} {Generative agents: Interactive simulacra of human behavior}.
\newblock \emph{ArXiv}, abs/2304.03442.

\bibitem[{Peskov and Cheng(2020)}]{peskov2020takes}
Denis Peskov and Benny Cheng. 2020.
\newblock It takes two to lie: One to lie, and one to listen.
\newblock In \emph{Proceedings of ACL}.

\bibitem[{Potthast et~al.(2017)Potthast, Kiesel, Reinartz, Bevendorff, and Stein}]{potthast2017stylometric}
Martin Potthast, Johannes Kiesel, Kevin Reinartz, Janek Bevendorff, and Benno Stein. 2017.
\newblock A stylometric inquiry into hyperpartisan and fake news.
\newblock \emph{arXiv preprint arXiv:1702.05638}.

\bibitem[{Radford et~al.(2023)Radford, Kim, Xu, Brockman, McLeavey, and Sutskever}]{radford2023robust}
Alec Radford, Jong~Wook Kim, Tao Xu, Greg Brockman, Christine McLeavey, and Ilya Sutskever. 2023.
\newblock Robust speech recognition via large-scale weak supervision.
\newblock In \emph{International Conference on Machine Learning}, pages 28492--28518. PMLR.

\bibitem[{Rapoza(2021)}]{Rapoza_2021}
Kenneth Rapoza. 2021.
\newblock \href {https://www.forbes.com/sites/kenrapoza/2017/02/26/can-fake-news-impact-the-stock-market/?sh=13be87f82fac} {Can “fake news” impact the stock market?}

\bibitem[{Serra-Garcia and Gneezy(2021)}]{serra2021mistakes}
Marta Serra-Garcia and Uri Gneezy. 2021.
\newblock Mistakes, overconfidence, and the effect of sharing on detecting lies.
\newblock \emph{American Economic Review}, 111(10):3160--3183.

\bibitem[{Shu et~al.(2017)Shu, Sliva, Wang, Tang, and Liu}]{shu2017fake}
Kai Shu, Amy Sliva, Suhang Wang, Jiliang Tang, and Huan Liu. 2017.
\newblock Fake news detection on social media: A data mining perspective.
\newblock \emph{ACM SIGKDD explorations newsletter}, 19(1):22--36.

\bibitem[{Siegler(1966)}]{siegler1966lying}
Frederick~A Siegler. 1966.
\newblock Lying.
\newblock \emph{American Philosophical Quarterly}, 3(2):128--136.

\bibitem[{Soldner et~al.(2019)Soldner, P{\'e}rez-Rosas, and Mihalcea}]{soldner2019box}
Felix Soldner, Ver{\'o}nica P{\'e}rez-Rosas, and Rada Mihalcea. 2019.
\newblock Box of lies: Multimodal deception detection in dialogues.
\newblock In \emph{Proceedings of the 2019 Conference of the North American Chapter of the Association for Computational Linguistics: Human Language Technologies, Volume 1 (Long and Short Papers)}, pages 1768--1777.

\bibitem[{Tafjord et~al.(2022)Tafjord, Dalvi, and Clark}]{Tafjord2022EntailerAQ}
Oyvind Tafjord, Bhavana Dalvi, and Peter Clark. 2022.
\newblock \href {https://api.semanticscholar.org/CorpusID:253097865} {Entailer: Answering questions with faithful and truthful chains of reasoning}.
\newblock In \emph{Conference on Empirical Methods in Natural Language Processing}.

\bibitem[{Tetterton and Warren(2005)}]{tetterton2005using}
Veronica~S Tetterton and Amye~R Warren. 2005.
\newblock Using witness confidence can impair the ability to detect deception.
\newblock \emph{Criminal Justice and Behavior}, 32(4):433--451.

\bibitem[{Touvron et~al.(2023)Touvron, Martin, Stone, Albert, Almahairi, Babaei, Bashlykov, Batra, Bhargava, Bhosale, Bikel, Blecher, Ferrer, Chen, Cucurull, Esiobu, Fernandes, Fu, Fu, Fuller, Gao, Goswami, Goyal, Hartshorn, Hosseini, Hou, Inan, Kardas, Kerkez, Khabsa, Kloumann, Korenev, Koura, Lachaux, Lavril, Lee, Liskovich, Lu, Mao, Martinet, Mihaylov, Mishra, Molybog, Nie, Poulton, Reizenstein, Rungta, Saladi, Schelten, Silva, Smith, Subramanian, Tan, Tang, Taylor, Williams, Kuan, Xu, Yan, Zarov, Zhang, Fan, Kambadur, Narang, Rodriguez, Stojnic, Edunov, and Scialom}]{Touvron2023Llama2O}
Hugo Touvron, Louis Martin, Kevin~R. Stone, Peter Albert, Amjad Almahairi, Yasmine Babaei, Nikolay Bashlykov, Soumya Batra, Prajjwal Bhargava, Shruti Bhosale, Daniel~M. Bikel, Lukas Blecher, Cristian~Cant{\'o}n Ferrer, Moya Chen, Guillem Cucurull, David Esiobu, Jude Fernandes, Jeremy Fu, Wenyin Fu, Brian Fuller, Cynthia Gao, Vedanuj Goswami, Naman Goyal, Anthony~S. Hartshorn, Saghar Hosseini, Rui Hou, Hakan Inan, Marcin Kardas, Viktor Kerkez, Madian Khabsa, Isabel~M. Kloumann, A.~V. Korenev, Punit~Singh Koura, Marie-Anne Lachaux, Thibaut Lavril, Jenya Lee, Diana Liskovich, Yinghai Lu, Yuning Mao, Xavier Martinet, Todor Mihaylov, Pushkar Mishra, Igor Molybog, Yixin Nie, Andrew Poulton, Jeremy Reizenstein, Rashi Rungta, Kalyan Saladi, Alan Schelten, Ruan Silva, Eric~Michael Smith, R.~Subramanian, Xia Tan, Binh Tang, Ross Taylor, Adina Williams, Jian~Xiang Kuan, Puxin Xu, Zhengxu Yan, Iliyan Zarov, Yuchen Zhang, Angela Fan, Melanie Kambadur, Sharan Narang, Aurelien Rodriguez, Robert Stojnic, Sergey Edunov, and
  Thomas Scialom. 2023.
\newblock \href {https://api.semanticscholar.org/CorpusID:259950998} {Llama 2: Open foundation and fine-tuned chat models}.
\newblock \emph{ArXiv}, abs/2307.09288.

\bibitem[{Van Der~Zee et~al.(2022)Van Der~Zee, Poppe, Havrileck, and Baillon}]{van2022personal}
Sophie Van Der~Zee, Ronald Poppe, Alice Havrileck, and Aur{\'e}lien Baillon. 2022.
\newblock A personal model of trumpery: linguistic deception detection in a real-world high-stakes setting.
\newblock \emph{Psychological science}, 33(1):3--17.

\bibitem[{Wang et~al.(2010)Wang, Spezio, and Camerer}]{wang2010pinocchio}
Joseph Tao-yi Wang, Michael Spezio, and Colin~F Camerer. 2010.
\newblock Pinocchio's pupil: using eyetracking and pupil dilation to understand truth telling and deception in sender-receiver games.
\newblock \emph{American economic review}, 100(3):984--1007.

\bibitem[{Wang et~al.(2023)Wang, Wei, Schuurmans, Le, Chi, Narang, Chowdhery, and Zhou}]{DBLP:conf/iclr/0002WSLCNCZ23}
Xuezhi Wang, Jason Wei, Dale Schuurmans, Quoc~V. Le, Ed~H. Chi, Sharan Narang, Aakanksha Chowdhery, and Denny Zhou. 2023.
\newblock \href {https://openreview.net/pdf?id=1PL1NIMMrw} {Self-consistency improves chain of thought reasoning in language models}.
\newblock In \emph{ICLR}. OpenReview.net.

\bibitem[{Wittenberg et~al.(2021)Wittenberg, Tappin, Berinsky, and Rand}]{wittenberg2021minimal}
Chloe Wittenberg, Ben~M Tappin, Adam~J Berinsky, and David~G Rand. 2021.
\newblock The (minimal) persuasive advantage of political video over text.
\newblock \emph{Proceedings of the National Academy of Sciences}, 118(47):e2114388118.

\bibitem[{Zuckerman et~al.(1981)Zuckerman, DePaulo, and Rosenthal}]{zuckerman1981verbal}
Miron Zuckerman, Bella~M DePaulo, and Robert Rosenthal. 1981.
\newblock Verbal and nonverbal communication of deception.
\newblock In \emph{Advances in experimental social psychology}, volume~14, pages 1--59. Elsevier.

\end{thebibliography}

\appendix

\section{Experiments and Results}
\paragraph{Base Models} For GPT-3.5 and GPT-4\footnote{\url{https://platform.openai.com/docs/models}}, we provide both the system and the user prompts. For GPT-3, there is only one prompt; hence the system and user prompts from above will be merged into one.  

\begin{figure*}[!h]
\small
\begin{framed}
System: You are an expert detective. \\

User: You are provided with a situation of deliberate misrepresentation. There are three contestants, Number One, Number Two, and Number Three, all claiming to be the same person. However, there is only one of them who is actually the person they are all claiming to be. There is an affidavit that includes truthful information about the claimed person. The affidavit is publicly available. However, the affidavit does not contain all the truth about the person, and it is possible that new information is obtained in the conversation that is actually true.\\

There is a judge who cross-questions all three contestants to uncover the truth. All the contestants will get monetary rewards if they can deceive the judge. However, the contestant, who is the real person, is sworn to speak truthfully. But they will most often utter half-truths to deceive the judges.\\

The name of the claimed person is given as Name:\\
The affidavit of the claimed person is given as Affidavit:\\
The conversation (in a question-answer form) between the judge and the contestants is given as Conversations:\\
If a question starts with addressing a particular contestant (e.g., Number One), then that question and all the following questions are addressed to that contestant unless a different contestant (e.g., Number Two) is explicitly addressed. \\ 

Based on the affidavit and the conversation, predict the contestant who is not an imposter. First, generate your rationale behind your prediction. Then, write \#\#\# followed by the single option from {Number One, Number Two, Number Three} as the answer. \\

<Input Name, Affidavit, Conversation>\\

Answer:

\end{framed}
\caption{System and User Prompts for Base Models}
\label{fig:system_prompt}
\end{figure*}

\paragraph{Bottleneck Models} We have four bottleneck prompts for entailment, ambiguity, overconfidence, and half-truths. 

\begin{figure*}[!h]
\small
\begin{framed}
We will analyze the conversation snippet by snippet in the order it happened originally. For every snippet:\\ 

\textbf{Entailment}: Cross-check the answer with the affidavit (you can verify the information from your factual knowledge). Use your judgment to generate a verdict that the answer(s) make the addressed contestant a likely imposter or likely the true person.\\

\textbf{Ambiguity}: A random or unrelated answer may indicate an imposter as they might be making up the facts. Based on this, generate a verdict that the answer(s) make the addressed contestant a likely imposter or likely the true person.\\

\textbf{Overconfidence}: An overconfident answer may indicate an imposter as they will go overboard to prove themselves as real. Based on this, generate a verdict that the answer(s) make the addressed contestant a likely imposter or likely the true person.\\

\textbf{Half-truths}: The real person is sworn to speak truthfully. Hence, in order to deceive, they may speak half-truths. Based on this, generate a verdict that the answer(s) make the addressed contestant a likely imposter or likely the true person.
\end{framed}
\caption{Bottleneck Prompt}
\label{fig:bottleneck_prompt}
\end{figure*}

\paragraph{Complete qualitative examples} \Cref{fig:1,fig:2,fig:3} contain complete illustrations of the three qualitative examples discussed in the main paper.

\begin{figure*}[!h]
\small
\begin{framed}
\textbf{Correct Prediction:} Number One\\
\textbf{Model Prediction:} Correct\\
\textbf{Judges Predictions:} All Wrong\\

\textbf{Model Explanation:}\\
Based on the affidavit and the conversation, here's my analysis:\\
Number One: This contestant gave short, incomplete but true answers. For example, when asked about the boat she took to England, she said she didn't go by boat because she gets seasick. This is a half-truth as she didn't specify how she actually got to England. When asked about what's on her mind after 21 hours in the water, she said she didn't remember anything. This is a generic answer that could be true. She also correctly named the two most important department stores in Toronto.\\
Number Two: This contestant gave long, detailed answers. For example, when asked about the grease used by channel swimmers, she specified axle grease. She also correctly answered the question about the governor general of Canada. However, she was unsure about the name of the famous hotel in Montreal, which could indicate that she is an imposter.\\
Number Three: This contestant gave a random answer when asked about the time it took Gertrude Ederle to swim the English Channel. She also gave a detailed answer about the food she eats when distance swimming, which could indicate that she is an imposter. However, she was truthful when she said she was afraid of her legs getting tired and of eels in the water.\\
Based on this analysis, I would rank the contestants in the following order of likelihood of not being an imposter: 1. Number One 2. Number Two 3. Number Three

\end{framed}
\caption{Qualitative example}
\label{fig:1}
\end{figure*}

\begin{figure*}[!h]
\small
\begin{framed}
\textbf{Correct Prediction:} Number One\\
\textbf{Model Prediction:} Correct\\
\textbf{Judges Predictions:} 2 Correct, 2 Wrong\\

\textbf{Model Explanation:}\\
Based on the affidavit and the conversation, we can make the following observations:\\
1. Number One: This contestant gave generic and incomplete answers, which is a characteristic of the real person according to the hints. For example, when asked about poultry pathology, they gave a very basic answer. They also struggled with the question about the size limit for a toy poodle, which could indicate unfamiliarity with the American measurement system, consistent with the affidavit stating that Maria is Brazilian. However, they failed to answer the question about candling an egg, which is a basic veterinary practice.\\
2. Number Two: This contestant gave detailed and confident answers, which is a characteristic of an imposter according to the hints. They knew the correct number of toes on a poodle and the correct term for a French poodle. They also knew the correct degree for a doctor of veterinary medicine. However, they incorrectly stated that there is only one variety of poodle, which is incorrect as there are three varieties: Toy, Miniature, and Standard.\\
3. Number Three: This contestant gave some correct answers, such as the life expectancy of a chicken and the meaning of "saudades". However, they also gave some incorrect and vague answers. For example, they failed to correctly name the three sizes of poodles and suggested that the cure for a common chicken disease is to break the chicken\'s neck, which is incorrect and unprofessional.\\
Based on these observations, the ranking from most likely to least likely to be the real Maria Angela Magalaich Dolovera is as follows:\\
1. Number One: Despite their failure to answer one question, their overall performance was most consistent with the hints about the real person's behavior.\\
2. Number Three: Although they gave some correct answers, their incorrect and vague answers suggest that they may be an imposter.\\
3. Number Two: Their detailed and confident answers, along with their incorrect statement about the varieties of poodles, strongly suggest that they are an imposter.

\end{framed}
\caption{Qualitative example}
\label{fig:2}
\end{figure*}

\begin{figure*}[!h]
\small
\begin{framed}
\textbf{Correct Prediction:} Number Three\\
\textbf{Model Prediction:} Wrong\\
\textbf{Judges Predictions:} All Correct\\

\textbf{Model Explanation:}\\
Based on the affidavit and the conversation, here's my analysis:\\
Number One: He seems to be truthful and not overly detailed in his responses. He doesn't know what a double clutch is, which is a bit suspicious for a driving champion, but it's possible he never encountered it. His answer about the gas station address is generic but could be true. He correctly identifies the octagon-shaped road sign as a stop sign, which is basic knowledge for a driver.\\
Number Two: He seems to be very detailed in his responses, which could indicate he's an imposter. He doesn't know what a double clutch is either, but he correctly identifies the diamond-shaped road sign as a hazard and explains the meaning of a broken line on the road. He also provides a specific address for his gas station, which could be a made-up detail.\\
Number Three: He gives a random answer about parking near a hydrant, which could indicate he's making up facts. He correctly identifies the square-shaped road sign as providing information about the road. His answer about double clutching is detailed and could be true, but it's also a bit overconfident. He provides a specific address for his gas station, which could be a made-up detail.\\
Based on these observations, I would rank the contestants in the following order of likelihood of not being an imposter: 1. Number One 2. Number Three 3. Number Two
\end{framed}
\caption{Qualitative example}
\label{fig:3}
\end{figure*}

\end{document}